\documentclass{article}

\usepackage[nonatbib,preprint]{nips_2018}

\usepackage[utf8]{inputenc} 
\usepackage[T1]{fontenc}    
\usepackage{hyperref}       
\usepackage{url}            
\usepackage{booktabs}       
\usepackage{amsfonts}       
\usepackage{nicefrac}       
\usepackage{microtype}      

\usepackage{amsmath}
\usepackage{graphicx}

\title{Unsupervised Learning by Competing Hidden Units}

\author{
  Dmitry Krotov \\
  MIT-IBM Watson AI Lab \& IBM Research\\
  Institute for Advanced Study \\
  \texttt{krotov@ibm.com} \\
   \And
John Hopfield \\
Princeton Neuroscience Institute  \\
Princeton University \\
   \texttt{hopfield@princeton.edu} \\
}

\begin{document}

\maketitle

\begin{abstract}
  It is widely believed that the backpropagation algorithm is essential for learning good feature detectors in early layers of artificial neural networks, so that these detectors are useful for the task performed by the higher layers of that neural network. At the same time, the traditional form of backpropagation is biologically implausible. In the present paper we propose an unusual learning rule, which has a degree of biological plausibility, and which is motivated by Hebb's idea that change of the synapse strength should be local - i.e. should depend only on the activities of the pre and post synaptic neurons.  We design a learning algorithm that utilizes global inhibition in the hidden layer, and is capable of learning early feature detectors in a completely unsupervised way. These learned lower layer feature detectors can be used to train higher layer weights in a usual supervised way so that the performance of the full network is comparable to the performance of standard feedforward networks trained end-to-end with a backpropagation algorithm.  
\end{abstract}

\section{Introduction}
Supervised learning with backpropagation at its core works extremely well on an immense diversity of complicated tasks \cite{HintonBengioLeCun}.  Using conventional techniques, the earliest layers of neurons in deep neural networks learn connections that yield neuron's receptive fields qualitatively described as feature detectors.  In visual tasks, the resemblance of some of the features found by backpropagation in convolutional neural networks to the simple observed selectivities of the response of neurons in early visual processing areas in the brains of higher animals is intriguing \cite{Fergus}. For simplicity, we will always describe the task to be performed as a visual task, but none of our methods have any explicit limitation to vision. All the methods discussed below are pixel and color permutation invariant.  

In concept, the learning rule that shapes the early artificial neural network responses through supervised learning, e.g. backpropagation or stochastic gradient decent (SGD), and the learning rules which describe the development of the early front-end neural processing in neurobiology are unrelated. There are two conceptual reasons for this. First, in real biological neural networks, the neuron responses are tuned by a synapse-change procedure that is physically local, and thus describable by local mathematics. Consider the synaptic connection $W_{i j}$ between an input neuron $i$ and an output neuron $j$. In SGD training the alteration of $W_{i j}$ depends not only on the activities of neurons $i$ and $j$, but also on the labels and the activities of the neurons in higher layers of the neural network, which are not directly knowable from the activities of neurons $i$ and $j$. Thus, the learning rule is non-local, i.e. requires information about the state of other specific neurons in the network in addition to the two neurons that are connected by the given synapse. In biology, the synapse update depends on the activities of the presynaptic cell and the postsynaptic cell and perhaps on some global variables such  as how well the task was carried out (state of animal attention, arousal, fear, etc.), but not on the activities other specific neurons. 

  Second, higher animals require extensive sensory experience to tune the early (in the processing hierarchy) visual system into an adult system. This experience is believed to be predominantly observational, with few or no labels, so that there is no explicit task. The learning is said to be unsupervised.   By contrast, training a deep artificial neural network (ANN) with SGD requires a huge amount of labeled data. 

The central question that we investigate in this paper is the following: can good (i.e. useful) early layer representations be learned in an ANN context without supervision and using only a local ``biological'' synaptic plasticity rule? We propose a family of learning rules that have conceptual biological plausibility and make it possible to learn early representations that are as good as those found by end-to-end training with SGD on MNIST. On CIFAR-10 the performance of our algorithm is slightly poorer than end-to-end training, but it still outperforms previously published benchmarks on biological learning. We demonstrate these points by training the early layer representations using our algorithm, then freezing the weights in that layer and adding another layer on top of it (that is trained with labels using conventional methods of SGD) to perform the classification task. 

Backpropagation can also be used to generate useful early representations without using labeled data. A feed-forward network with a small bottleneck structure can be trained to carry out a self-mapping task on input patterns \cite{autoencoders}.  The bottleneck serves the role of the labels in providing information to be back-propagated to determine connection patterns in early processing layers.  The bottleneck generates virtual labels which are non-locally propagated backward.  This is very different from our system which requires no back-propagation of signals, receives all its information directly from forward-propagating signals, and has a local rule of synapse update.  

For completeness, we note that diverse efforts have been made to describe a procedure equivalent to backpropagation for training an ANN without grossly violating the ``rules'' of neurobiology.  In principle this is possible, but only with considerable addition to the present simple network structure and/or invoking particular special aspects of neuronal biophysics.   The direction of such efforts is well summarized in \cite{Bengio_video, Hinton_video, DTP, Bengio, Lillicrap, Ororbia1, Ororbia2}. 

What is meant by ``biological plausibility'' for synapse change algorithms in ANN?  We take as fundamental the idea coming from psychologists that the change of the efficacy of synapses is central to learning, and that the most important aspect of biological learning is that the coordinated activity of a presynaptic cell $i$ and a post-synaptic cell $j$ will produce changes in the synaptic efficacy of the synapse $W_{i j}$ between them \cite{Hebb}. In the years since Hebb's original proposal, a huge amount of neurobiology research has fleshed out this idea.   We now know that neurons make either excitatory or inhibitory outgoing synapses (but not both) \cite{Eccles}, that for excitatory synapses the change in efficacy can either increase or decrease depending on whether the coordinated activity of the post-synaptic cell is strong or weak (LTP=long term potentiation and LTD=long term depression) \cite{Koch}, that homeostatic processes regulate overall synaptic strength, that pre-post timings of neuronal action potentials can be very important (STDP) \cite{Gerstner}, that changes in synaptic efficacy may be quantized \cite{Petersen}, that there are somewhat different rules for changing the strength of excitatory and inhibitory synapses \cite{Luo}, etc.  For the  ANN purposes, we will subsume most such detail with four ideas:
\begin{itemize}
\item  The change of synapse strength during the learning process is proportional to the activity of the pre-synaptic cell and to a function of the activity of the post-synaptic cell.  Both LTP (hebbian learning) and LTD (anti-hebbian learning) are important.  The synapse update is locally determined.

\item Lateral inhibition between neurons within a layer, which makes the network not strictly feedforward, is responsible for developing a diversity of pattern selectivity across many cells within a layer.

\item The effect of limitations of the strength of a synapse and homeostatic processes will bound possible synaptic connection patterns. The dynamics of our weight-change algorithm expresses such a bound, so that eventually the input weight vector to any given neuron converges to lie on the surface of a (unit) sphere.

\item The normalization condition could emphasize large weights more than small ones. In order to achieve this flexibility we construct the (local) dynamics of the synapse development during learning so that the fixed points of this dynamics can be chosen to lie on a Lebesgue sphere with $p$-norm, for $p \geq 2$. 
\end{itemize}

\section{Mathematical framework}
\begin{figure}[h]
\begin{center}
\includegraphics[width = 0.5\linewidth]{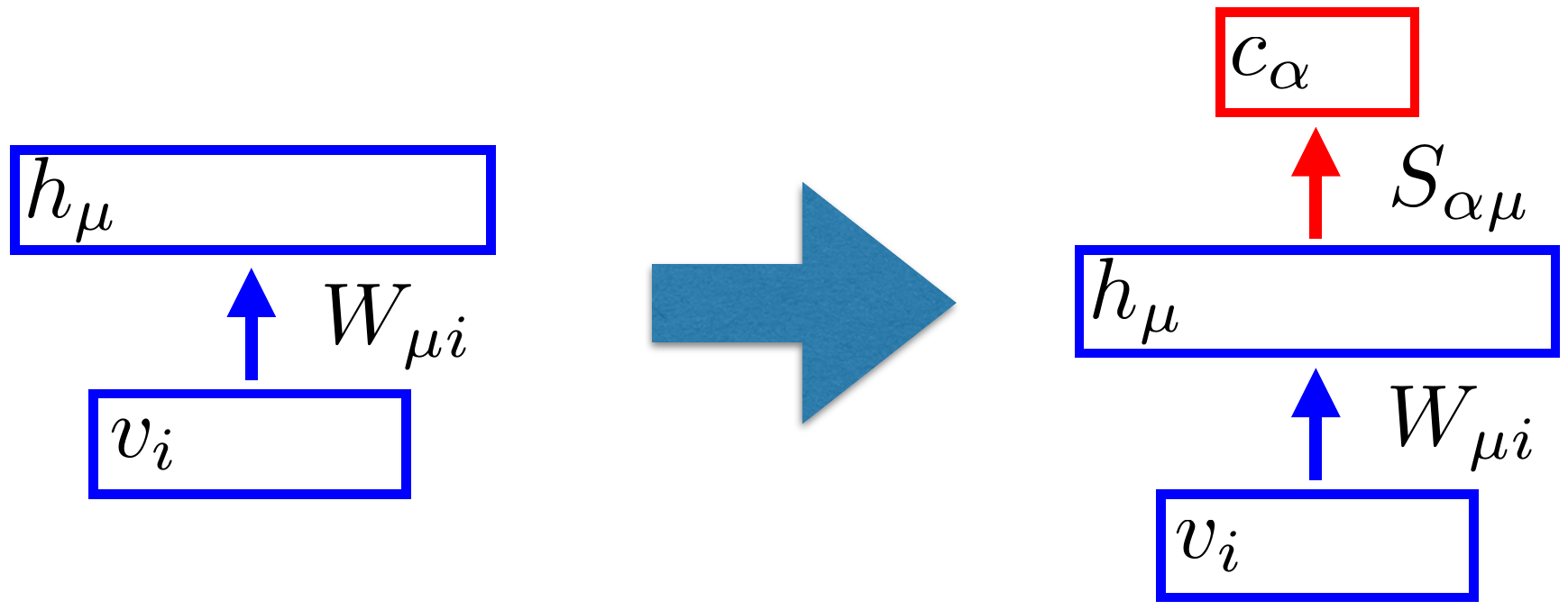}
\end{center}
\caption{\footnotesize{Lower layers of the neural network (wieghts $W_{\mu i}$) are trained using a biological learning algorithm. Once this phase is complete, the weights are plugged into a fully connected perceptron. The weights of the top layer $S_{\alpha\mu}$ are then learned using SGD in a supervised way.}}\label{logic}
\end{figure}

A standard feed-forward architecture is considered with visible neurons $v_i$, several hidden layers, and a layer of output neurons $c_\alpha$. We illustrate the logic with one hidden layer. In this case the forward pass is specified by the equations (summation over repeated indices $i$ and $\mu$ is assumed)
\begin{equation}
\begin{cases}
h_\mu=f(W_{\mu i} v_i)\\
c_\alpha= \tanh(S_{\alpha\mu} h_\mu)
\end{cases} \ \text{where}\ \ \ \ f(x)=\begin{cases}
x^n, \ \ \ \ x\geq0\\
0, \ \ \ \ \ \ x<0
\end{cases}
\label{network equations}
\end{equation}
The power of the activation function $n \geq 1$ is a hyperparameter of the model \cite{Krotov Hopfield 2016, Krotov Hopfield 2017}, $n=1$ corresponds to ReLU. The receptive fields of the hidden layer $W_{\mu i}$ are learned using our local unsupervised algorithm, described below, without any information about the labels. Once this unsupervised part of the training in complete, the second set of weights $S_{\alpha\mu}$ are learned using conventional SGD techniques. This is the only part of the training algorithm where labels are utilized. This logic is illustrated in Fig.\ref{logic}.  

It has been known since the work of Hubel and Wiesel that many neurons in the visual cortex are tuned to detect certain elementary patterns of activity in the visual stimulus. A biologically plausible mechanism for development of direction selectivity in the visual cortex was suggested by Bienenstock, Cooper, and Munro \cite{BCM}. The idea of BCM theory is that for a random sequence of input patterns a synapse is learning to differentiate between those stimuli that excite the postsynaptic neuron strongly and those stimuli that excite that neuron weakly. Learned BCM feature detectors cannot however be simply used as the lowest layer of a feedforward network so that the entire network is competitive to a network of the same size trained with backpropagation algorithm end-to-end. The main reason for this is that BCM is a theory of the development of the pattern selectivity of a single cell.  Highly specific pattern responses in BCM come about because there is a temporal {\it competition between patterns seeking to drive a single neuron}.  This competition is controlled by the dynamics of an adjustable learning threshold parameter. In our system, {\it the neurons compete with each other for patterns}, and there is no adjustable threshold. The competition is between neurons, not between patterns. When one neuron becomes tuned to some pattern of inputs, the within-layer lateral inhibition keeps other neurons from becoming selective to that same pattern.  Thus a layer of diverse early feature detectors can be learned in a completely unsupervised way without any labels. 

\subsection{Synaptic plasticity rule}
Consider the case of only one hidden unit. Then the matrix $W_{\mu i }$ becomes a vectors ${\bf W}_i$. It is convenient to define a metric and an inner product ($\delta_{ij}$ is Kronecker delta)
$$
\langle {\bf X}, {\bf Y}\rangle= \sum\limits_{i,j} g_{i j} {\bf X}_i{\bf Y}_j, \ \ \ \ \text{with} \ \ \ \ g_{ij}= |{\bf W}_i|^{p-2} \delta_{i j}	
$$
where $p\geq1$ is the parameter defining Lebesgue $p$-norm. The plasticity rule that we study can then be written as 
\begin{equation}
\tau_L \frac{d{\bf W}_i}{dt} = g\big(Q\big) \Big(R^p {\bf v}_i - \big\langle{\bf W},{\bf v}\big\rangle {\bf W}_i \Big),\ \ \ \text{where} \ \ \ Q=\frac{\big\langle{\bf W},{\bf v}\big\rangle\ \ \ }{\big\langle{\bf W},{\bf W}\big\rangle^{\frac{p-1}{p}}}\label{learning rule}
\end{equation}
The constant $\tau_L$ defines the time scale of the learning dynamics.  It should be larger than the time scale of presentation of an individual training example, as well as the time scale of evolution of individual neurons, which is defined below. The function $g(Q)$ is a non-linear learning activation function that is discussed below; ${\bf v}_i$ are visible neurons or training examples. During the unsupervised phase of the training the weights are initialized from a standard normal distribution. Then a sequence of training examples is presented one at a time and the weights are updated according to (\ref{learning rule}). It can be shown that as $t\rightarrow\infty$ the weights $W_i$ converge to a sphere of radius $R$, defined using $L^p$ norm.
$$
|W_1|^p + |W_2|^p + ... + |W_N|^p = R^p
$$ 
In order to see this consider the time derivative of the $p$-norm of the weight vector
$$
\tau_L \frac{d}{dt} \big\langle{\bf W},{\bf W}\big\rangle = p g(Q) \big\langle{\bf W},{\bf v}\big\rangle\Big[ R^p - \big\langle{\bf W},{\bf W}\big\rangle \Big]
$$
Provided that $g(Q) \big\langle{\bf W},{\bf v}\big\rangle\geq 0$, if the $p$-length of vector ${\bf W}$ is less than $R$ its length increases on the dynamics, if it is greater than $R$ its length decreases. Thus, although the training procedure starts with random values of the synaptic connections, eventually these connections converge to a sphere. The positivity bound $g(Q) \big\langle{\bf W},{\bf v}\big\rangle\geq 0$ is satisfied for a broad class of activation functions, for example for any power $n$ in (\ref{network equations}). In practical applications, we will use this learning rule even in situations when the positivity constraint is violated. It turns out that if the violation is weak, which is justified by the small parameter $\Delta$ (see Fig.\ref{pipeline} below), the weights still converge to a sphere of radius $R$.

In situations when the network has more than one hidden unit, which are enumerated by index $\mu$, each vector ${\bf W}_{\mu}$ will have an external index, as will the inner product. In this case the fixed points of the plasticity dynamics are such that synapses of each hidden unit converge to their own unit vector on a sphere. 

It can also be shown that plasticity rule (\ref{learning rule}) has a Lyapunov function (see appendix A) 
\begin{equation}
L= \sum\limits_{\mu=1}^K G\Bigg[ \frac{\big\langle{\bf W}_{\mu},{\bf v}\big\rangle_{\mu}\ \ \ }{\big\langle{\bf W}_{\mu},{\bf W}_{\mu}\big\rangle^{\frac{p-1}{p}}_{\mu}} \Bigg], \ \ \ \text{where}\ \ \ G'(Q) = g(Q)\label{Lyapunov function}
\end{equation}
that monotonically increases on the dynamical trajectory of synapses. Lastly, if one works on the ordinary Euclidean sphere with $p=2$ and the function $g(Q)=Q$ is linear, our learning rule reduces to the famous Oja rule \cite{Oja}. In the following the radius of the sphere is set to $R=1$. 

\subsection{A biologically inspired learning algorithm}
A good set of weights in the neural network should be such that different hidden units detect different features of the data. Neither BCM algorithm, nor the plasticity rule (\ref{learning rule}) address this issue of differential selectivity. In order to do this, we introduce a neural network with global inhibition between the hidden neurons and dynamical equations
\begin{equation}
\tau\frac{dh_\mu}{dt} = I_\mu - w_{\text{inh}}\sum\limits_{\nu\neq\mu} r(h_\nu) - h_\mu, \ \ \ \ \text{where}\ \ \ \ I_\mu=\langle{\bf W}_\mu, {\bf v}\rangle\label{dynamics}
 \end{equation} 
In this equation the activities of the hidden neurons are denoted by $h_\mu$, $r(h_\mu)=\max(h_\mu,0)$ are the corresponding firing rates (we use a ReLU), $I_\mu$ is the input current from the visible layer, $w_{\text{inh}}$ is a parameter that defines the strength of the global inhibition, constant $\tau\ll \tau_L$ defines the dynamical time scale of individual neurons.
\begin{figure}[h]
\begin{center}
\includegraphics[width = 1.0\linewidth]{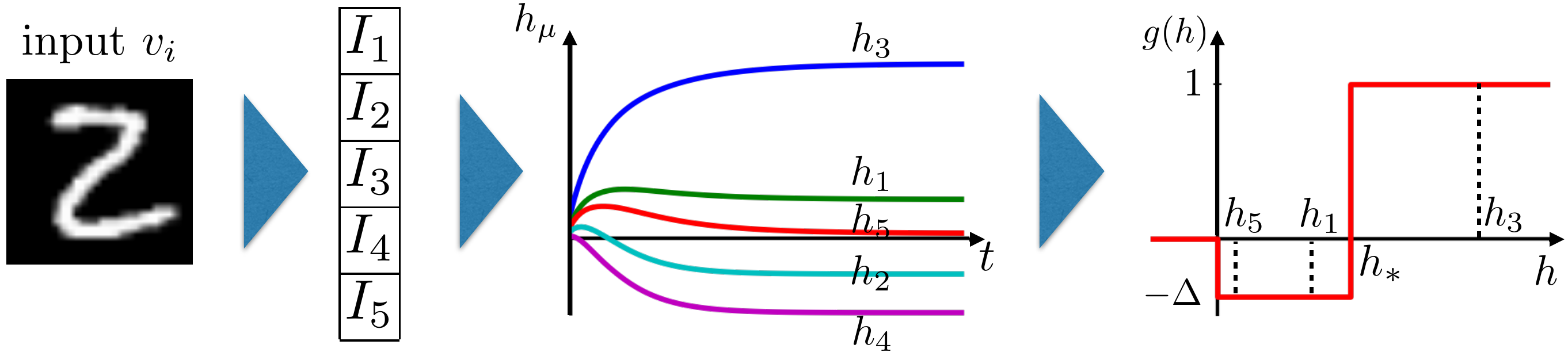}
\end{center}
\caption{\footnotesize{The pipeline of the training algorithm. Inputs $v_i$ are converted to a set of input currents $I_\mu$. They define the dynamics (\ref{dynamics}) that leads to the steady state activations of the hidden units. These activations are used to update the synapses using the learning rule (\ref{learning rule}). The learning activation function changes the sign at $h_*$, which separates the hebbian and anti-hebbian learning regimes.}}\label{pipeline}
\end{figure} 

The pipeline of the unsupervised part of the algorithm in shown in Fig.\ref{pipeline}. Using the weights $W_{\mu i}$ a raw input is converted into a set of input currents $I_\mu$. These currents drive the dynamics of hidden units, as is shown in Fig.\ref{pipeline}. The strength parameter of the global inhibition is set so that in the final state only a small fraction of hidden units have positive activity (in this case $h_3$, $h_1$, and $h_5$). The values of this steady state activities are then used as arguments in the non-linear learning activation function $g(h)$ in (\ref{learning rule}). This function implements temporal competition between the patterns, so that it is positive for the activities exceeding a threshold $h_*$, and negative for the activities in the range $0<h_\mu<h_*$. Activities that are below zero do not contribute to training. The intuitive idea behind this choice of the activation function is that the synapses of hidden units that are strongly driven are pushed towards the patterns that drive them, while the synapses of those hidden units that are driven slightly less are pushed away from these patterns. Given a random temporal sequence of the input stimuli this creates a dynamic competition between the hidden units and results in the synaptic weights that are different for each hidden unit and specific to features of the data. The idea of having an activation function that is positive for activations above the threshold $h_*$, and negative below the threshold $h_*$ is inspired by the BCM theory \cite{BCM} and the existence of LTP and LTD \cite{Koch}. 

\section{A fast AI implementation}
We view the algorithm presented in the previous section as a conceptual idea of how it might be possible to learn good early layer representations given the biological constraints of the sensory cortex. An advantage of the presented algorithm, compared to SGD, is that it is unsupervised. The question then arises, whether this algorithm might be valuable from the AI perspective if we forget about its biological motivation. 

From the AI point of view the main drawback of the presented algorithm is that it is slow. There are two reasons for this. First, it is an online algorithm so that training examples are presented one at a time, unlike SGD where training examples can be presented in minibatches. Second, for any training example one has to wait until the set of hidden units reaches a steady state. This requires numerically solving Eq (\ref{dynamics}), which is time consuming. 

 An approximation of this computational algorithm has been found which circumvents these two drawbacks and works extremely well in practice. First, instead of solving dynamical equations we will use the currents as a proxy for ranking of the final activities. Given that ranking the unit that responds the most to a given training example is pushed towards that example with activation $g=1$. The unit that is second (or more generally $k$-th) in ranking is pushed away from the training example with activation $g=-\Delta$. This heuristic significantly speeds up the algorithm. Second, training examples can then be organized in minibatches, so that the ranking is done for the entire minibatch and the weight updates that result from the learning rule (\ref{learning rule}) are averaged over all examples in the minibatch.

\section{Testing the model}
The presented algorithm was tested on the MNIST and CIFAR-10 datasets. All the tasks that are discussed below are pixel, and (in the case of CIFAR-10) color permutation invariant. The hyperparameters are reported in Appendix B, supplementary. 
\subsection*{MNIST}
A standard training set of 60000 examples was randomly split into 50000 examples training set and a 10000 examples validation set that was used for tuning the hyperparameters. After the hyperparameters were fixed, the training and the validation sets were combined to train the final model on 60000 examples. The final performance of the models was evaluated on the standard held out test set of 10000 examples. 

\begin{figure}[h]
\begin{center}
\includegraphics[width = 1.0\linewidth]{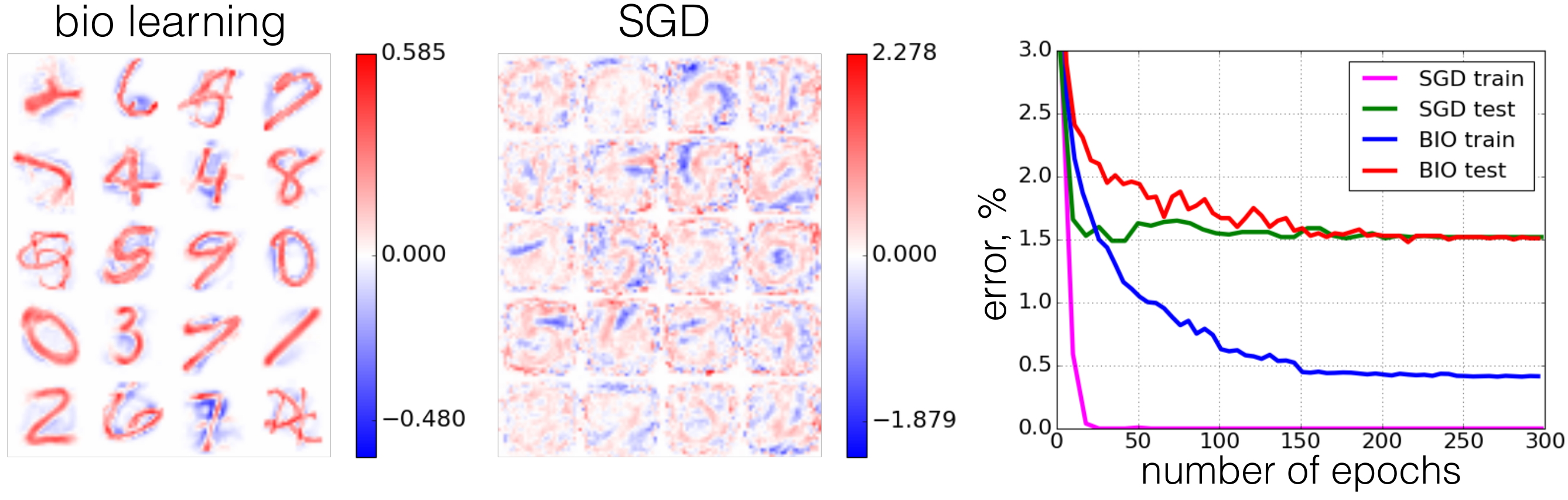}
\end{center}
\caption{\footnotesize{(Left) The weights learned by the network using the unsupervised learning algorithm, 20 randomly chosen feature detectors out of 2000 are shown. (Middle) The weights learned by the network trained end-to-end with SGD algorithm, 20 randomly chosen feature detectors out of 2000 are shown. (Right) Error rate on the training and test sets as training progresses for the ``biological'' algorithm and for the standard SGD.}}\label{bio_vs_SGD}
\end{figure}

In the first set of experiments a network of 2000 hidden units was trained using the biological algorithm to find the weights $W_{\mu i}$. The optimal value of the Lebesgue norm parameter $p=4$ was used. The initial values for weights were initialized from the standard normal distribution. As training progresses the weights eventually converge to the surface of a unit sphere. Final learned weights, connected to 20 randomly chosen hidden units are shown in Fig.\ref{bio_vs_SGD} left panel.
After the unsupervised phase of the training was complete, the learned weights $W_{\mu i}$ were frozen and used in the network (\ref{network equations}). The second set of weights was learned using a standard SGD procedure described in appendix B, suppl. The decrease of the error on the training and the test sets during this second (supervised) phase of the training are shown in Fig.\ref{bio_vs_SGD} right panel, blue and red curves. The performance of this ``biologically'' trained network was compared with the performance of the feedforward network of the same size ($784 \rightarrow 2000 \rightarrow 10$) trained end-to-end with SGD starting from random weights (training and test errors are shown in Fig.\ref{bio_vs_SGD} right panel, magenta and green curves). Randomly chosen 20 feature detectors learned by the standard SGD network are shown in Fig.\ref{bio_vs_SGD} middle panel. 

The network trained with SGD end-to-end reaches the well known benchmarks: training error$=0\%$, test error$\approx1.5\%$. Training error of the ``biological'' network is $0.41\%$. The most surprising aspect of this plot is that the test error of the ``biological'' algorithm, which is $1.52\%$, is the same as the test error of the network trained end-to-end with SGD. This is significantly better than the previously published benchmarks on other biologically inspired algorithms, for example $2\!-\!3\%$ in \cite{Bengio}, $1.96\%$ in \cite{Sacramento}, $1.94\%$ in \cite{DTP}, $5\%$ in \cite{Diehl}. When comparing these numbers it should also be noted that the algorithms of \cite{Bengio, Sacramento, DTP} are using the information about the labels all the way during the training. In contrast, our algorithm learns the first layer representations in a completely unsupervised way. Thus it demonstrates a better performance in spite of solving a more challenging task.  

It is instructive to examine the weights (Fig.\ref{bio_vs_SGD}, left) learned by the unsupervised phase of the ``biological'' algorithm. The color code uses the white color to represent zero weights, the red color to represent positive weights, and the blue color to represent negative weights. Training examples $v_i$ are normalized so that they are always positive $v_i\geq0$. The weights associated with each hidden unit live on a unit sphere, i.e. satisfy the constraint $\sum\limits_{i=1}^{784}|W_{\mu i}|^p=1$, which is the fixed point of the learning rule (\ref{learning rule}).  Although some of the learned feature detectors resemble certain training examples, they are not simply copies of the individual training data points. The easiest way to see this is to notice that the largest positive weight among the 20 shown feature detectors, $0.585$, is almost as big as the largest (in absolute value) negative weight $-0.48$, while the training examples are always positive. Thus the learned feature detectors encode both where the ink is in the data, and where the ink is not. Mathematically, these negative weights arise because of the anti-hebbian piece in the learning rule proportional to $-\Delta$. 

Another important aspect is that the ``biological'' network is not doing a simple template matching in order to classify the data. For example feature detectors 1 and 5 (counting is left to right top to bottom) encode sub-digit features - slanted line and a corner. Feature detector 4 ``votes'' for class 7 and against class 0. Feature detectors 3 and 20 are activated by classes 4, 5, and 2,4 respectively. Thus, the weights learned by the network encode a distributed over multiple hidden units representation of the data. The same conclusion could be obtained by examining the patterns of activations of hidden units by a given input image. This representation is however very different from the representation learned by the SGD network as is clear from comparison of the left and the middle panels in Fig.\ref{bio_vs_SGD}. 

Lastly, the structure of MNIST images is such that pixels on the periphery of the images are almost always in the off state, except for a few examples in the training set. Thus it is unnatural for a good learning algorithm to learn large weights associated with those uninformative pixels. However, as is clear from the middle panel in Fig.\ref{bio_vs_SGD}, the SGD network does develop full scale weights associated with those pixels in almost all the feature detectors (except for the pixels that are off in strictly all the 60000 training examples). Thus those pixels can be used to generate adversarial images that fool the neural network. In contrast, our algorithm learns weights that are equal to zero at the periphery of the images, which is more natural.      

It is instructive to learn the first layer weights $W_{\mu i}$ for the optimal values of the hyperparameters, then freeze those weights, and then learn the second layer weights $S_{\alpha\mu}$ for different values of power $n$ in (\ref {network equations}). 
\begin{figure}[h]
\begin{center}
\includegraphics[width = 1.0\linewidth]{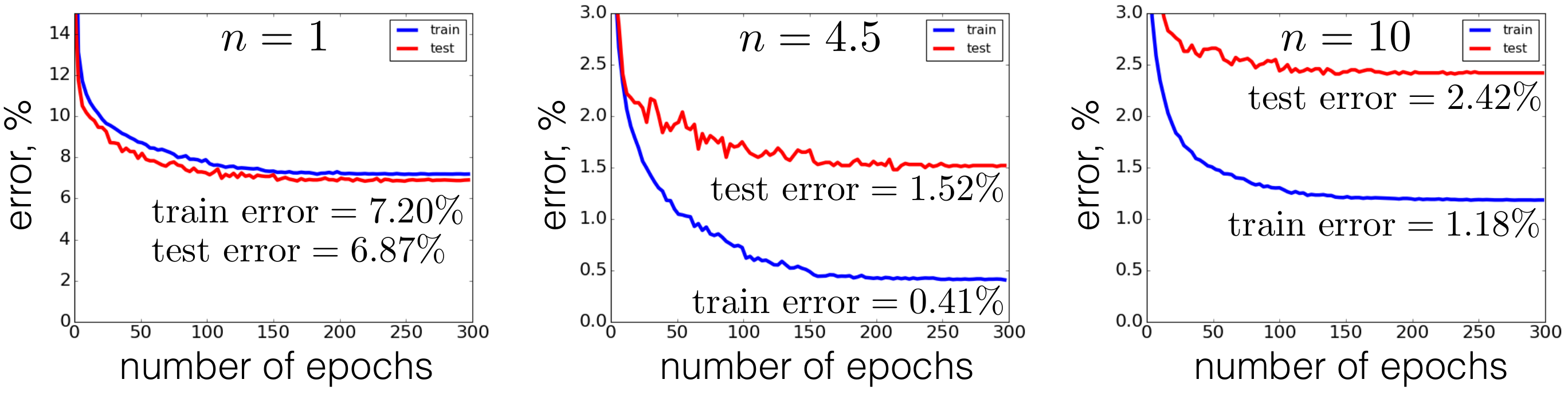}
\end{center}
\caption{\footnotesize{Error on the training set (blue) and the test set (red) for three networks with different powers of the activation function $f(x)$. All networks were trained for 500 epochs with the schedule of the learning rate annealing described in Appendix B.}}\label{effect_of_n}
\end{figure} 
The hyperparameters associated with the second supervised phase of the training are optimized for each value of power $n$. The results are shown in Fig.\ref{effect_of_n}. There is an optimal value of power $n\approx 4.5$ for which generalization error and training error are the smallest. The error increases for both smaller and larger values of power $n$. This indicates that given the weights, which are learned by the unsupervised phase of the algorithm, the architecture of higher layers should be tuned so that the network could take most advantage of those unsupervised weights.  

\subsection*{CIFAR-10}
A standard training set of 50000 examples was randomly split into 45000 examples training set and 5000 examples validation set. After tuning the hyperparameters on the validation set, the final models were trained on the entire 50000 data points and evaluated on the held-out test set of 10000 examples. No preprocessing of the data was used, except that each input image was normalized to be a unit vector in the $32*32*3=3072$ dimensional space. 

In analogy with MNIST the performance of two networks was compared. One trained using the ``biological'' algorithm in two phases: unsupervised training of the $3072\rightarrow 2000$ network, followed by the supervised training of the top layer $2000\rightarrow10$. The other one $3072\rightarrow2000\rightarrow10$ trained with SGD end-to-end. 
\begin{figure}[h]
\begin{center}
\includegraphics[width = 1.0\linewidth]{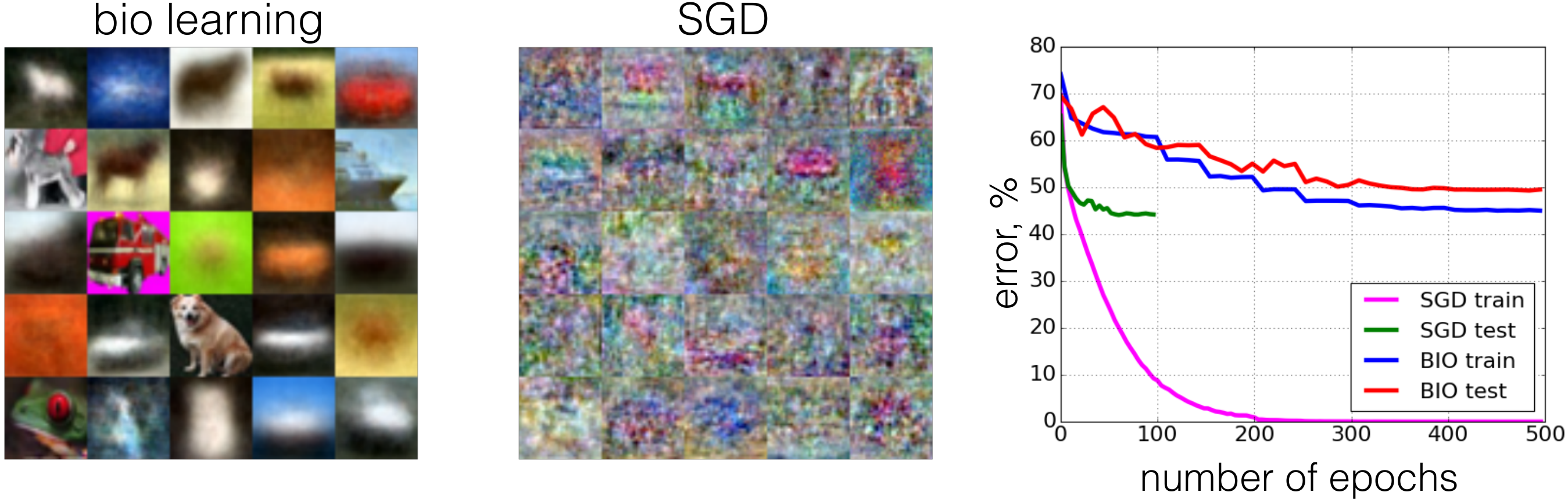}
\end{center}
\caption{\footnotesize{(Left) The weights learned by the network using the unsupervised learning algorithm, 25 randomly chosen feature detectors out of 2000 are shown. (Middle) The weights learned by the network trained end-to-end with SGD algorithm, 25 randomly chosen feature detectors out of 2000 are shown. (Right) Error rate on the training and test sets as training progresses for the biological algorithm and for the standard end-to-end network.}}\label{bio_vs_SGD_CIFAR}
\end{figure} 
The results are shown in Fig.\ref{bio_vs_SGD_CIFAR}. The optimal duration of SGD training for the conventional network, see Appendix B for details, was 100 epochs, which gave the error on the test set $44.74\%$. At that point the error on the training set is still non-zero. To demonstrate that the capacity of the SGD network is sufficient to fit all the training examples, we kept training it until the training error reached $0\%$. The ``biological'' network achieves  training error $44.95\%$ and test error $49.25\%$. Unfortunately, previously published benchmarks for biologically inspired training algorithms are very scarce on CIFAR. The only benchmark that we found is \cite{DTP}, which reports $49.29\%$. It is worth emphasizing again that the algorithm of \cite{DTP} is supervised all the way throughout training and for this reason solves a simpler task compared to our algorithm that learns the first set of weights in an entirely unsupervised way. In Fig.\ref{bio_vs_SGD_CIFAR} the weights of 25 randomly selected hidden units learned by the unsupervised phase of the algorithm are shown. They represent a diverse collection of features and prototypes of the training images. These weights have large negative elements, and for this reason are not just copies of the training examples, which are all positive. The classification decision of the full ``biological'' network is done in a distributed way involving votes of multiple hidden units. This distributed representation is very different, however, from the end-to-end feature detectors shown in the middle panel. Lastly, although the network is dealing with the pixel-color permutation invariant problem, the unsupervised algorithm discovers the continuity of color in the data. This is very different from the SGD feature detectors, which look~speckled.

\section{Discussion and conclusions} 
Historically, neurobiology has inspired much research on using various plasticity rules to learn useful representations from the data. This line of research chiefly disappeared after 2014 because of the success of deep neural networks trained with backpropagation on complicated tasks like ImageNet. This has led to the opinion that neurobiology-inspired plasticity rules are computationally inferior to networks trained end-to-end, and that supervision is crucial for learning useful early layer representations from the data. By consequence, the amount of attention given to exploring the diversity of possible biologically-inspired learning rules, in the present era of large data sets and fast computers, has been rather limited. Our paper  challenges this opinion by describing an unsupervised learning algorithm that demonstrates a very good performance on MNIST and CIFAR. The core of the algorithm is a local learning rule, that incorporates both LTP and LTD types of plasticity and a network motif with global inhibition in the hidden layer. 
 
 In the present paper all the experiments were done on a network with one hidden layer. The proposed unsupervised algorithm, however, is iterative in nature. This means that after a one layer representation is learned, it can be used to generate the codes for the input images.  These codes can be used to train another layer of weights using exactly the same unsupervised algorithm. There are many possibilities of how one could organize those additional layers, since they do not have to be fully connected. This line of research will be described in a follow-up paper. 
 
 Lastly, the SGD training of the top layer was used in this paper to assess the quality of the representations learned by the unsupervised phase of our algorithm. This does not invalidate the biological plausibility of the entire algorithm, since SGD in one layer can be written as a local synaptic plasticity rule involving only a pre and post synaptic cell activities. Thus, it complies with the locality requirement that we took as fundamental. 
  	                     
\section*{Appendix A}               
Below we prove that (\ref{Lyapunov function}) monotonically increases on the dynamics (\ref{learning rule}). Consider one row of the matrix $W_{\mu i}$. The temporal derivative of (\ref{Lyapunov function}) is then given by
\begin{equation}
\begin{split}
\tau_L \frac{dL}{dt} = \sum\limits_{\mu=1}^K \frac{(p-1)g(Q_\mu)}{\langle{\bf W}_\mu, {\bf W}_\mu \rangle^{\frac{p-1}{p}+1}}\Big[\tau_L \Big\langle\frac{d{\bf W}_\mu}{dt}, {\bf v}\Big\rangle \Big\langle{\bf W}_\mu,{\bf W}_\mu\Big\rangle - \tau_L \Big\langle\frac{d {\bf W}_\mu}{dt}, {\bf W}_\mu\Big\rangle \Big\langle{\bf W}_\mu, {\bf v}\Big\rangle   \Big] =\\ =
\sum\limits_{\mu=1}^K \frac{(p-1)g(Q_\mu)^2 R^p}{\langle{\bf W}_\mu, {\bf W}_\mu \rangle^{\frac{p-1}{p}+1}}\Big[ \Big\langle{\bf W}_\mu,{\bf W}_\mu\Big\rangle\langle{\bf v}, {\bf v}\Big\rangle - \Big\langle{\bf W}_\mu, {\bf v}\Big\rangle^2\Big] \geq 0	
\end{split}
\end{equation}
The last expression is positive due to the Cauchy–Schwarz inequality.

\section*{Appendix B}
Unsupervised part of the training for MNIST was done for the following optimal parameters: Lebesgue norm $p=4$, anti-hebbian learning parameter $\Delta=0.4$. Training was done using minibatches of size $100$ examples for 100 epochs. Learning rate linearly decreased from the maximal value $0.04$ at the first epoch to $0$ at the last epoch. The learning rate here is defined as the numerical coefficient that is used to multiply the right hand side of (\ref{learning rule}) to obtain the increment in weights. 

Supervised part of the training was done using the loss function (labels $t_\alpha$ are one-hot encoded vectors of $N_c=10$ units of $\pm 1$)
$$
C=\sum\limits_{\text{examples}}\sum\limits_{\alpha=1}^{N_c} |c_\alpha - t_\alpha|^m
$$
Adam optimizer was used to minimize the loss function during 300 epochs with the following schedule of learning rate change: 0.001 for the first $100$ epochs, after that the learning rate decreased every 50 epochs as $0.0005$, $0.0001$, $0.00005$, $0.00001$.  Minibatch size was $100$, $m=10$. Power of the activation function was $n=4.5$ for the experiments shown in Fig.\ref{bio_vs_SGD}, and as specified in Fig.\ref{effect_of_n} for that figure. 
  
For unsupervised part of CIFAR-10 experiments, a similar setting was used with $p=4$, $\Delta=0.3$. Training was done for 100 epochs with minibatches of size $100$. Learning rate linearly decreased from $0.02$ to $0$. 

For the supervised part of ``biological'' network Adam optimizer was used with minibatch size $10$, $m=6$, $n=10$ for $500$ epochs. The learning rate schedule: $0.004$ for $100$ epochs, then each $50$ epochs the learning rate changed as $0.002$, $0.001$, $0.0005$, $0.0002$, $0.0001$, $0.00005$, $0.00002$, $0.00001$. 

For the supervised part of SGD network Adam optimizer was used with minibatch size $100$ and $m=4$, $n=1$ for $100$ epochs. Learning rate: $0.004$ for $50$ epochs, then $0.001$ for another $50$ epochs.

\section*{Acknowledgements}
We thank Shiyu Chang, David Cox, Guy Gur-Ari, Arnold Levine and Sijia Liu for useful discussions. The work of DK at IAS was partly supported by the Starr Foundation.


\begin{thebibliography}{99}
\bibitem{HintonBengioLeCun} LeCun, Y., Bengio, Y. and Hinton, G., 2015. Deep learning. Nature, 521(7553), p.436.

\bibitem{Fergus} Zeiler, M.D. and Fergus, R., 2014, September. Visualizing and understanding convolutional networks. In European conference on computer vision (pp. 818-833). Springer, Cham.

\bibitem{autoencoders} Hinton, G.E. and Salakhutdinov, R.R., 2006. Reducing the dimensionality of data with neural networks. science, 313(5786), pp.504-507.

\bibitem{Bengio_video} Bengio, Y., 2017.  Deep Learning and Backprop in the Brain (Online video). Available: https://www.youtube.com/watch?v=FhRW77rZUS8.

\bibitem{Hinton_video} Hinton, G., 2016. Can Sensory Cortex Do Backpropagation? (Online video). Available: https://www.youtube.com/watch?v=cBLk5baHbZ8

\bibitem{DTP} Lee, D.H., Zhang, S., Fischer, A. and Bengio, Y., 2015, September. Difference target propagation. In Joint european conference on machine learning and knowledge discovery in databases (pp. 498-515). Springer, Cham.

\bibitem{Bengio} Scellier, B. and Bengio, Y., 2017. Equilibrium propagation: Bridging the gap between energy-based models and backpropagation. Frontiers in computational neuroscience, 11, p.24.

\bibitem{Lillicrap} Lillicrap, T.P., Cownden, D., Tweed, D.B. and Akerman, C.J., 2016. Random synaptic feedback weights support error backpropagation for deep learning. Nature communications, 7, p.13276.

\bibitem{Ororbia1} Ororbia, A.G., Mali, A., Kifer, D. and Giles, C.L., 2018. Conducting credit assignment by aligning local representations. arXiv preprint arXiv:1803.01834.

\bibitem{Ororbia2} Ororbia, A.G. and Mali, A., 2019, July. Biologically motivated algorithms for propagating local target representations. In Proceedings of the AAAI Conference on Artificial Intelligence (Vol. 33, pp. 4651-4658).

\bibitem{Hebb} Hebb, D.O.,1949. The organization of behavior: A neurophysiological approach, pp.62

\bibitem{Eccles} Eccles, J., 1976. From Electrical to Chemical Transmission in the Central Nervous System. Notes and Records of the Royal Society of London, 30(2), pp.219-230.

\bibitem{Koch} Koch, C., 2004. Biophysics of computation: information processing in single neurons. Oxford university press.

\bibitem{Gerstner} Gerstner, W., Kistler, W.M., Naud, R. and Paninski, L., 2014. Neuronal dynamics: From single neurons to networks and models of cognition. Cambridge University Press.

\bibitem{Petersen} Petersen, C.C., Malenka, R.C., Nicoll, R.A. and Hopfield, J.J., 1998. All-or-none potentiation at CA3-CA1 synapses. Proceedings of the National Academy of Sciences, 95(8), pp.4732-4737.

\bibitem{Luo} Luo, L., 2015. Principles of neurobiology. Garland Science.

\bibitem{Krotov Hopfield 2016} Krotov, D. and Hopfield, J.J., 2016. Dense associative memory for pattern recognition. In Advances in Neural Information Processing Systems (pp. 1172-1180).

\bibitem{Krotov Hopfield 2017} Krotov, D. and Hopfield, J.J., 2017. Dense associative memory is robust to adversarial inputs. arXiv preprint arXiv:1701.00939.

\bibitem{BCM} Bienenstock, E.L., Cooper, L.N. and Munro, P.W., 1982. Theory for the development of neuron selectivity: orientation specificity and binocular interaction in visual cortex. Journal of Neuroscience, 2(1), pp.32-48.

\bibitem{Oja} Oja, E., 1982. Simplified neuron model as a principal component analyzer. Journal of mathematical biology, 15(3), pp.267-273.


\bibitem{Diehl} Diehl, P.U. and Cook, M., 2015. Unsupervised learning of digit recognition using spike-timing-dependent plasticity. Frontiers in computational neuroscience, 9, p.99.

\bibitem{Sacramento} Sacramento, J., Costa, R.P., Bengio, Y. and Senn, W., 2017. Dendritic error backpropagation in deep cortical microcircuits. arXiv preprint arXiv:1801.00062.


\end{thebibliography}
\end{document}